\pdfoutput=1
\documentclass[journal]{IEEEtran}
\usepackage{amsmath,amsfonts}
\usepackage{algorithmic}
\usepackage{array}
\usepackage{textcomp}
\usepackage{stfloats}
\usepackage{url}
\usepackage{verbatim}
\usepackage{graphicx}
\def\BibTeX{{\rm B\kern-.05em{\sc i\kern-.025em b}\kern-.08em
		T\kern-.1667em\lower.7ex\hbox{E}\kern-.125emX}}
\usepackage{balance}
\usepackage{verbatim}
\usepackage{amsfonts}
\usepackage{amssymb}
\usepackage{stfloats}
\usepackage{cite}
\usepackage{soul}
\usepackage{setspace}
\usepackage{graphicx}
\usepackage{psfrag}
\usepackage{amsmath} 
\usepackage{array}
\usepackage{epstopdf}
\usepackage{authblk}
\usepackage{graphicx} 
\usepackage{amsthm} 
\usepackage{lipsum}
\usepackage{verbatim} 
\usepackage{authblk}
\usepackage{mathtools}
\usepackage{cuted}
\usepackage{bm}
\usepackage[lined,boxed,ruled]{algorithm2e}
\usepackage{booktabs}
\usepackage{subfigure}
\usepackage[font=small]{caption} 

\usepackage{ifthen}

\usepackage[usenames]{color}

\newtheorem{proposition}{Proposition}[section]

\usepackage{hyperref}

\begin{document}
	\title{Generative Feature Imputing --- A Technique for Error-resilient Semantic Communication}
	\author{
		{Jianhao~Huang, Qunsong~Zeng, Hongyang Du, ~\IEEEmembership{Member,~IEEE},  and Kaibin~Huang ~\IEEEmembership{Fellow,~IEEE} }
		\thanks{
	{J. Huang, Q. Zeng, H. Du, and K. Huang are with the Department of Electrical and Electronic Engineering, The University of Hong Kong, Hong Kong.  Emails: \{jianhaoh,   duhy, huangkb\}@hku.hk, qszeng@eee.hku.hk. Corresponding author: K. Huang. }
}
}

	\maketitle
	\thispagestyle{empty}

\begin{abstract}

Semantic communication (SemCom) has emerged as a promising paradigm for achieving unprecedented communication efficiency in sixth-generation (6G) networks by leveraging artificial intelligence (AI) to extract and transmit the underlying meanings of source data. However, deploying SemCom over digital systems presents new challenges, particularly in ensuring robustness against transmission errors that may distort semantically critical content. To address this issue, this paper proposes a novel framework, termed generative feature imputing, which comprises three key techniques. First, we introduce a spatial-error-concentration packetization strategy that spatially concentrates feature distortions by encoding feature elements based on their channel mappings—a property crucial for both the effectiveness and reduced complexity of the subsequent techniques. Second, building on this strategy, we propose a generative feature imputing method that utilizes a diffusion model to efficiently reconstruct missing features caused by packet losses. Finally, we develop a semantic-aware power allocation scheme that enables unequal error protection by allocating transmission power according to the semantic importance of each packet. Experimental results demonstrate that the proposed framework outperforms conventional approaches, such as Deep Joint Source-Channel Coding (DJSCC) and JPEG2000, under block fading conditions, achieving higher semantic accuracy and lower Learned Perceptual Image Patch Similarity (LPIPS) scores.

\end{abstract}


	\section{Introduction}
	
	The sixth-generation (6G) wireless networks promise to support a broad range of emerging applications, such as immersive internet-of-things (IoT), multimedia streaming, and augmented reality, which necessitate ultra-high rates and reliability, and low latency \cite{saad2019vision,11121577,lin2024efficient,10453386,10529950}. However, as dictated by Shannon’s information theory, these objectives are in conflict with each other given limited radio resources \cite{polyanskiy2010channel}.  Traditional communication systems mitigate this conflict mainly by enhancing transmission capacity through expanded bandwidth, higher power, or larger antenna arrays \cite{Tse2005, 7930469}. These approaches, however,  result in ever-increasing system complexity and resource consumption. As a paradigm-shifting solution, semantic communication (SemCom) leverages artificial intelligence (AI) to extract and transmit critical features of source data, thereby dramatically reducing communication overhead \cite{zhang2022toward,10175391,10287247,li2023fundamental}.  However, implementing SemCom over 6G networks  presents critical challenges, including integrating AI computation with modern communication protocols.
	
	Modern communication systems ensure reliable end-to-end (E2E) transmission through rigorous bit-error control, targeting error rate below $10^{-5}$
	in 5G standards \cite{shafi20175g}. The emphasis on accurate bit transmission stems from the  separate architecture of source and channel codings. On one hand, source coding techniques, such as entropy coding \cite{witten1987arithmetic} and vector quantization \cite{quantization}, exploit statistical dependencies within data to achieve high compression efficiency. On the other hand, channel coding provides all  bits with a uniform level of protection  against transmission errors. Although this modular design offers  simplicity and practicability, it is highly susceptible to channel noises. Specifically, decoding errors introduced by noisy channels propagate through source decoders due to inherent data dependencies, leading to potentially abrupt performance loss—a phenomenon termed the \emph{cliff effect} \cite{10845799}. 
	
	Recent physical-layer advancements aim to mitigate the performance degradation by either  enhancing channel capacity (e.g., via multi-antenna systems and expanded spectrum resources \cite{7930469}) or improving error control (e.g., through stronger error-correction codes \cite{choukroun2022error}, adaptive modulation \cite{950343}, or packet retransmission \cite{6888474}). However, these methods increase computational complexity and encounter challenges of practical scalability. Meanwhile, source coding approaches like Multiple Description Coding (MDC) \cite{1369698} and JPEG2000 \cite{1037027} have sought to reduce dependencies within encoded streams to improve error resilience. For instance, JPEG2000 employs layered compression to partition images into independently encoded blocks, thereby reducing error propagation across blocks \cite{1037027}. However, these strategies  sacrifice coding efficiency due to redundant data (e.g., block headers) and the lack of semantic-aware representations of source data. These limitations necessitate a paradigm shift from accurate bits delivery to meaning transmissions that are much more error-resilient and efficient.

Empowered by AI, SemCom  has emerged as a promising paradigm for enhancing error resilience in data transmissions. One widely recognized solution is merging the traditionally separate branches of source and channel coding to enable an E2E coded transmission system. 
By leveraging the deep neural networks (DNNs), a so-called deep joint source-channel coding (DJSCC) scheme directly maps the source data (e.g., image \cite{10024766,dai2022nonlinear,bourtsoulatze2019deep}, text \cite{9398576}) into a feature space for transmissions over analog channels.  By jointly optimizing source  and channel codings in an E2E manner, DJSCC mitigates the cliff effect and achieves superior robustness against channel noise  compared to traditional separation-based schemes.
However, the analog  outputs of the DJSCC encoder are incompatible with modern digital communication infrastructures, which are built on digital modulation and channel coding schemes.  To address this issue, 
researchers have introduced schemes that quantize the continuous output vectors from DNNs into finite sets of constellation points, establishing a digital DJSCC framework \cite{9998051,bo2023joint}.   However, these methods are still unable to leverage the existing designs of powerful digital codes.

	To ensure compatibility with existing digital communication systems, recent efforts have focused on integrating semantic analysis into modular-based source-channel coding frameworks \cite{he2023rate,liu2024ofdm}. By leveraging AI algorithms, the transmitter identifies the most semantically critical information in the data, enabling channel encoders to apply unequal error protection (UEP) to the bit streams against channel noise and fading. This UEP strategy enhances task-oriented robustness by tolerating higher bit error rates (BER) in less critical data segments. Specifically, rate-adaptive coding mechanisms have been proposed to dynamically allocate channel rates across multimodal data streams based on their semantic importance \cite{he2023rate}. To further improve allocation efficiency, deep reinforcement learning (DRL) algorithms have been developed to optimize sub-carrier and bit allocation by modeling interdependencies between semantic features and downstream tasks \cite{liu2024ofdm}. 

The most recent breakthrough in SemCom leverages the remarkable power of generative AI (GenAI) to impute lost data using received low-dimensional prompts, thereby attaining unprecedented communication efficiencies  \cite{rombach2022high,lugmayr2022repaint,qiao2025token, lin2023pushing}.
 Specifically, diffusion models can synthesize missing image pixels through iterative refinement of noisy inputs, enabling high-fidelity reconstruction \cite{rombach2022high, lugmayr2022repaint}. This capability allows systems to prioritize transmission of critical semantic features while delegating the recovery of missing details to generative models at the receiver \cite{huang2025visual}. However, the computational demands of such approaches remain prohibitive for real-time applications. Processing high-dimensional data (e.g., ultra-high-definition images exceeding 3840×2160 pixels) requires extensive iterative sampling steps, posing significant latency challenges. 
To mitigate this issue, a promising solution is to reduce data redundancy and capture critical features through semantic encoding \cite{10845799,huang2025visual}. 
As generative SemCom is still in its nascent stage, there lacks systematic design approaches that jointly consider semantic coding, UEP, and generative decoding to maximize error resilience while reining in computational complexity.

In this paper, we fill the void by proposing a novel framework of generative feature imputing to mitigate the impact of packet errors in digital SemCom systems. The framework employs semantic encoding for feature extraction and block channel coding for error protection.  By designing the framework, we aim to address two key challenges: (1) the cliff effect caused by random packet errors, and (2) the high computational complexity associated with GenAI-based data imputation. To this end, the proposed framework incorporates three core techniques, summarized as follows.
	
\begin{itemize}
	\item \textbf{Feature Error Concentration}: To mitigate the cliff effect, we propose a spatial-error-concentration packetization strategy that encodes extracted feature elements into packets according to their channel mappings, thereby ensuring that feature distortions caused by packet errors are spatially concentrated.  This strategy is feasible based on the fact that the feature elements across different channel mappings typically capture only localized spatial information of the image.  When combined with the following imputation technique, the spatial concentration enabled by the proposed strategy will significantly enhance the system’s error resilience. 
	
	\item \textbf{Generative Feature Imputing}: Building upon the preceding strategy, we further propose a generative feature imputing method to reconstruct missing feature elements due to packet loss.  The novelty of this approach lies primarily in identifying spatial error locations using packet loss information obtained from channel decoding.  Given these locations and the successfully decoded feature elements, the diffusion model iteratively recovers the missing features from Gaussian noise, guided by the learned feature distribution.  In contrast to conventional generative approaches that operate in high-dimensional data domains, the proposed feature imputing method leverages spatially concentrated errors to significantly reduce computational complexity, thereby making it well-suited for real-time applications. 
	
	\item \textbf{Semantic-aware Power Allocation}: To enhance the effectiveness of generative feature imputing, we propose the advanced technique of  semantic-aware power allocation (PA)  for UEP. First, we employ class activation mapping (CAM) to estimate the semantic importance of transmitted packets. Based on this importance modeling, we optimize the PA across orthogonal frequency bandwidths by minimizing the weighted sum of packet error rates. This ensures that semantically critical features receive a  higher level of protection, while less important features can be reconstructed via generative imputation.

\end{itemize}

Experimental results demonstrate that the proposed GenAI-enabled coding scheme effectively mitigates the cliff effect in digital SemCom systems and exhibits improved robustness against block fading compared to conventional benchmarking schemes such as DJSCC. Due to the absence of joint coding across all features, the proposed scheme might exhibit reduced performance as measured using data-level metrics, e.g., lower Peak Signal-to-Noise Ratio (PSNR) values than other source coding schemes. However, it achieves lower Learned Perceptual Image Patch Similarity (LPIPS) scores and higher semantic accuracy, which are more important in the context of SemCom.

The remainder of this paper is organized as follows. The system model of the error-resilient SemCom is introduced in Section II. The proposed generative feature imputing is presented in Section III. The semantic-aware PA scheme is proposed in Section IV. Experimental results are presented in Section V, followed by concluding remarks in Section VI.

 \begin{figure*}[t]
	\normalsize
	\setlength{\abovecaptionskip}{+0.3cm}
	\setlength{\belowcaptionskip}{-0.1cm}
	\centering
	\includegraphics[width=0.90\linewidth]{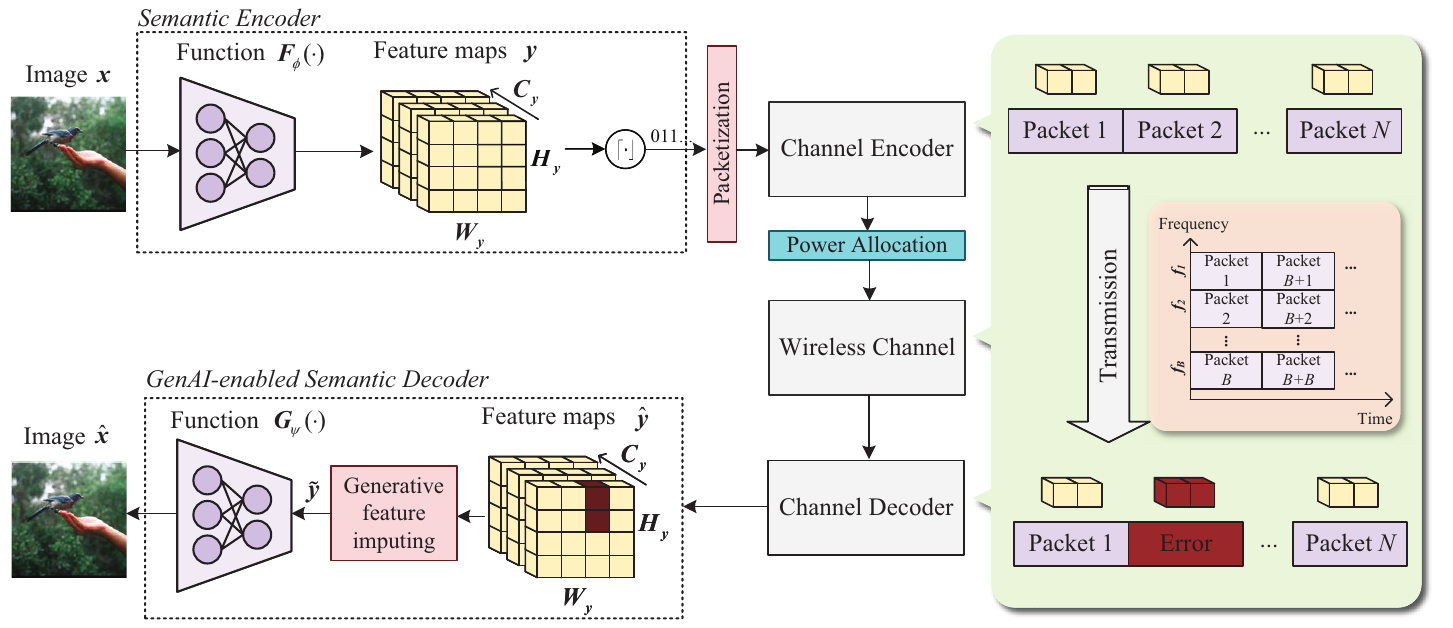}
	\captionsetup{justification=justified}
	\caption{The error-resilient digital SemCom system}
	\label{architecture}
\end{figure*}

\section{Overview of the Digital SemCom System}
The proposed framework of generative feature imputing targets a digital SemCom system. In this section, we introduce the system model operations and  performance metrics.
\subsection{System Blocks and Operations}

Consider an E2E SemCom system for wireless image transmission, where image samples are lossily compressed by DNNs and transmitted to the receiver for recovery. The recovered images can be used for various applications, e.g., human vision and machine tasks. To be compatible with the current communication protocols, we employ a separation-based architecture design, which divides the coding process by two modules: semantic encoding and and  block channel coding, as illustrated in the sequel. 
\subsubsection{Semantic Encoder}
Consider a colored image sample $\bm{x}\in \mathbb{R}^{W \times H \times 3}$ with $W$ and $H$ being the width and height, respectively. As shown in Fig. \ref{architecture}, the image sample, $\bm{x}$, is input into a convolutional neural network (CNN) based encoder function, denoted by $F_{\phi}(\cdot)$, to extract its continuous feature $\bm{y} \in \mathbb{R}^{W_y\times H_y \times C_y}$ with $W_y$, $H_y$,  and $C_y$ being the width, height, and number of feature channels, respectively. In other words, 
\begin{align}
	\bm{y}=F_{\phi}(\bm{x}),
\end{align}
where $\phi$ denotes the  parameter set of the CNN-based function. Let $y_{ijc}$ denotes the $(i,j,c)$-th element of feature $\bm{y}$.  To avoid error propagation, we  utilize the suboptimal but low-complexity uniform scalar quantization, instead of variable-length source coding,  to encode feature $\bm{y}$.    For each feature element $y_{ijc}$, we have a same quantization codebook of a total $2^{R}$, which is denoted by $\mathcal{U}=\{u_{1},\cdots,u_{2^{R}}\}$. Then, the quantization resolution is calculated as 
\begin{align}
	\triangle=\frac{U_{max}-U_{min}}{2^{R}-1},
\end{align}
where $U_{max}$ and $U_{min}$ are the estimated maximal and minimal values of feature elements. We denote the quantization operation as $\mathcal{Q}_{\triangle}(\cdot)$, which employs the element-wise quantization on the feature $\bm{y}$. The quantizer output is given by
\begin{align} \label{quantizaiotn}
	\bm{y}_{Q}=\mathcal{Q}_{\triangle}(\bm{y})=\{\left\lceil {y}_{ijc} \right\rfloor\} \in \mathcal{U}^{W_y\times H_y \times C_y},
\end{align}
where the operation $\left\lceil y \right\rfloor=u_i$ for $y \in \left[ u_{i}-\frac{\triangle}{2}, u_{i}+\frac{\triangle}{2}\right)$.  Accordingly, the bit stream for encoding  image  sample $\bm{x}$ is represented as $\bm{b} \in \{0,1\}^{L}$, where the source rate $L$ is calculated by $L=R  W_y  H_y  C_y$.

\subsubsection{Finite Blocklength Transmissions} The bit stream $\bm{b}$ is divided into $N$ packets of equal block length for transmission, where each packet contains $K$ information bits. For each packet, we employ $(K, D)$ block channel coding with $D$ being the block length.  Hence, the channel rate is given by $R_c=\frac{K}{D}$. 
It is important to note that the bits in $\bm{b}$ are not encoded into packets randomly or sequentially. Instead, they follow a \emph{structured packetization order} to enhance error-resilience performance, which will be  detailed in Section III.

The encoded packets are allocated to $B$ orthogonal sub-carriers for transmissions with $B<N$. During transmissions, we consider an independent and identically distributed (i.i.d.) block fading channel, where channel coefficients remain constant for each packet but vary across packets. The received signals for the $i$-th packet, denoted as $\bm{r}_i \in \mathbb{C}^{D}$, is given by
\begin{equation}
	\bm{r}_i = h_i \sqrt{P_i}  \bm{c}_i + \bm{n}_i,
\end{equation}
where $\bm{c}_i$ is the codeword of channel coding with a unit power, $h_i \in \mathbb{C}$ is the channel gain for the $i$-th block,  and $\bm{n}_i$ is the additive white Gaussian noise (AWGN)  with zero mean and variance $\sigma^2\bm{I}$. Here, we consider the channel state information at both the transmitter and receiver,   and thus power control can be applied. The transmitted power satisfies the average power constraint, i.e., $\frac{1}{B}\sum_{l=1}^{B}P_i\leq P_{ave}$, where $P_{ave}$ is the  power budget.
The received signal-to-noise ratio (SNR) of each packet is given by
\begin{align}
	\gamma_i=\frac{|h_i|^2P_i}{\sigma^2}. 
\end{align}

After receiving from the transmitter, the receiver decodes the information bits of each packet.  According to the finite block-length information theory \cite{polyanskiy2010channel}, the decoding error for the $i$-th packet can be approximated by 
\begin{align} \label{rho}
	\rho_{i}=Q\left(\sqrt{\frac{D}{V(\gamma_i)}}(C(\gamma_i)-R_c) \log 2 \right),
\end{align}
where $Q(\cdot)$ denotes the Q-function \cite{polyanskiy2010channel} defined as $Q(x)=\frac{1}{\sqrt{2\pi}} \int_{x}^{\infty} \exp(-\frac{t^2}{2}) dt$, $C(\gamma_i)$ is the channel capacity denoted by 
\begin{align}
	C(\gamma_i)=\log_2(1+\gamma_i),
\end{align}
and $V(\gamma_i)$ is the channel dispersion \cite{polyanskiy2010channel} expressed as
\begin{align} \label{disper}
	V(\gamma_i)=\frac{\gamma_i(2+\gamma_i)}{(1+\gamma_i)^2}.
\end{align}

Note that the derivation of the  error probability in \eqref{rho} is based on random coding, which is an ideal coding to describe the performance bound of the digital system \cite{polyanskiy2010channel}.  However, the proposed error-resilient system is also applicable for other practical channel codings, e.g., LDPC and modulations by applying their corresponding error probability relationship. 

\subsubsection{GenAI-enabled Semantic Decoder} After channel decoding, the decoded bits will be dequantized as the distorted feature $\hat{\bm{y}} \in \mathcal{U}^{W_y\times H_y \times C_y}$. Due to the effect of deep fading, some packets are not successfully decoded and thus erased. Since we apply the uniform quantization, only features in the failed packets are affected.  Denote the position set of erased features as $\mathbb{P}=\{(i,j,c)\}$, which is known  at the receiver\footnote{The receiver can determine which packets are not successfully transmitted by leveraging error detection mechanisms such as cyclic redundancy check (CRC), parity bits, or sequence numbers embedded in the packet headers \cite{clark1981error}.}.  Then, the erased feature elements are set as constant values, i.e.,
\begin{align}
	\hat{y}_{ijc}=k_{ijc}, \forall (i,j,c) \in \mathbb{P}.
\end{align}
The constant value $k_{ijc}$  provides a biased estimate and will be detailed in section III.   

Next, the distorted feature $\hat{\bm{y}}$ will be input into a generative feature imputing function, denoted by $M_{\theta}(\cdot)$, to generate the features distorted by the packet errors.  The generator output can be expressed as 
\begin{align}
	\tilde{\bm{y}}=M_{\theta}(\hat{\bm{y}}),
\end{align}
where $\theta$ is the parameter set of the generative function. Finally, the generated feature $\tilde{\bm{y}}$ is fed into a decoder function $G_{\psi}(\cdot)$ to output the recovered image, i.e.,
\begin{align}
	\hat{\bm{x}}=G_{\psi}(\tilde{\bm{y}}), 
\end{align}
where $\psi$ denotes the parameter set of the decoder function.

\subsection{Performance Metrics}
To quantify the performance of the recovered image $\hat{\bm{x}}$, we consider both the visual metrics and semantic accuracy.

\subsubsection{Visual Metrics} We adopt two widely used visual metrics: PSNR and LPIPS \cite{zhang2018unreasonable}. The PSNR measures pixel-level fidelity between the recovered image $\hat{\bm{x}}$ and the reference image $\bm{x}$, which is defined as \cite{10845799}
\begin{equation}
	\text{PSNR}(\hat{\bm{x}}, \bm{x}) = 10 \log_{10} \left( \frac{\text{MAX}^2}{\text{MSE}(\hat{\bm{x}}, \bm{x})} \right),
\end{equation}
where $\text{MAX} = 255$, and the Mean Squared Error (MSE) is computed across all  pixels:
\begin{equation}
	\text{MSE}(\hat{\bm{x}}, \bm{x}) = \frac{1}{3WH} \sum_{i=1}^{W} \sum_{j=1}^{H} \sum_{c=1}^{3} \left( \hat{x}_{ijc} - x_{ijc} \right)^2.
	\end{equation}
	Here, $\hat{x}_{ijc}$ and $x_{ijc}$ denote the pixel values at position $(i,j,k)$ in the recovered and reference images, respectively.  The LPIPS metric  quantifies perceptual similarity using deep features extracted from a pre-trained neural network, which is computed as:
	\begin{align}
&\text{LPIPS}(\hat{\bm{x}}, \bm{x}) \\ \nonumber 
&= \sum_{\ell \in \mathcal{L}} \frac{1}{H_{\ell} W_{\ell} C_{\ell}} \sum_{h=1}^{H_{\ell}} \sum_{w=1}^{W_{\ell}} \sum_{c=1}^{C_{\ell}} \left\| \bm{F}^{(\ell)}(\hat{\bm{x}})_{hwc} - \bm{F}^{(\ell)}(\bm{x})_{hwc} \right\|_2^2,
\end{align}
where $\bm{F}^{(\ell)}(\cdot) \in \mathbb{R}^{H_{\ell}\times W_{\ell} \times C_{\ell}}$ represents the feature map at layer $\ell$ of the pre-trained network (e.g., AlexNet \cite{zhang2018unreasonable}), and $\mathcal{L}$ denotes the set of selected layers. Lower LPIPS values indicate better perceptual alignment with human vision.

\subsubsection{Semantic Accuracy}
To further examine the semantic accuracy of the recovered image, we evaluate the classification accuracy of $\hat{\bm{x}}$ using a pre-trained classifier, e.g., vision transformer \cite{dosovitskiy2021image}.  

 \begin{figure*}[t]
	\normalsize
	\setlength{\abovecaptionskip}{+0.3cm}
	\setlength{\belowcaptionskip}{-0.1cm}
	\centering
	\includegraphics[width=0.90\linewidth]{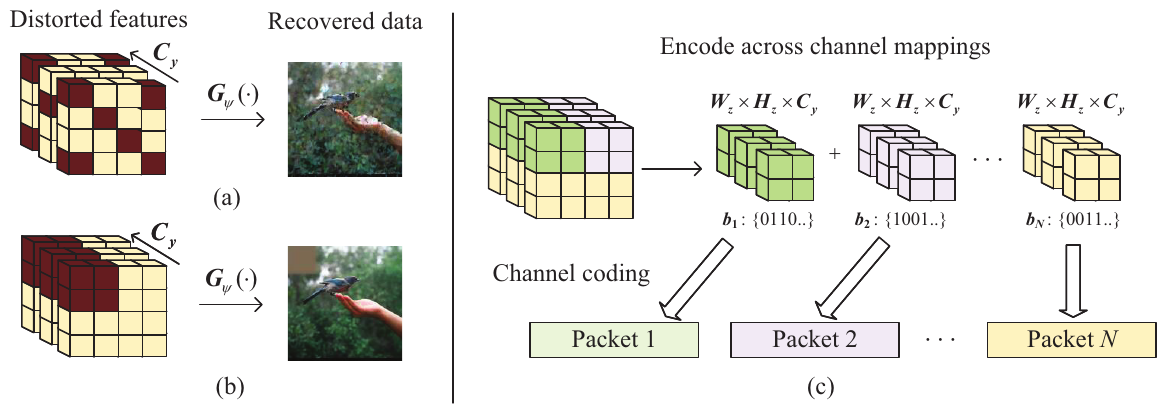}
	\captionsetup{justification=justified}
	\caption{Illustrations of the proposed spatial-error-concentration packetization method.  Sub-figures (a) and (b) represent the recovered images when features are  distorted randomly and over feature channels, respectively. The `red' features denote the ones affected by packet error. Sub-figure (c) illustrates the proposed packetization strategy according to channel mappings of features. }
	\label{error}
\end{figure*}

\section{ Generative Feature Imputing -- Basic Techniques}
In this section, we propose  the approach of the generative feature imputing to recover the lost features of the image.  First, we  present the  spatial-error-concentration packetization strategy to organize the features into ordered packets for transmissions. Then, based on the strategy,  we  introduce the diffusion-based generation model to recover the missing features due to packet loss. 

\subsection{Packetization with Spatial Error Concentration}

In this subsection, we propose the spatial-error-concentration  packetization strategy to manage the effect of packet errors on feature transmissions, which facilitates the subsequent design of  the generative imputing method. 

\subsubsection{Impact of Feature Error Position} First, we investigate the impact of packet error position on the image recovery, which motivates the design of the proposed packetization strategy. 
In the previous system model, we introduce the CNN-based encoder,  $F_{\phi}(\cdot)$, to extract feature maps $\bm{y} \in \mathbb{R}^{W_y \times H_y \times C_y}$ from the input image $\bm{x}$. The encoder $F_{\phi}(\cdot)$ leverages convolution layers to hierarchically capture spatial characteristics of $\bm{x}$, such as edges, textures, and object boundaries. This spatial sensitivity implies that errors in spatial dimensions ( along $W_y\times H_y$ axes) will proportionally affect the distortions on images. For instance, a perturbation localized to a spatial region $(w,h)$ in quantized feature $\bm{y}_{Q}$ can induce structured distortions in the reconstructed image, as the decoder amplifies spatial correlations during upsampling. To better illustrate this phenomenon, we let the feature elements be lost randomly, as depicted in Fig. \ref{error} (a). It is observed that although only a few errors occur at each feature channel, the recovered image is almost distorted at each pixel. 

Conversely, errors concentrated on the feature channels (i.e., along the $C_y$-axis of $\bm{y}_{Q}$) have a more localized impact. This asymmetry arises because individual channels often encode abstract visual attributes (e.g., color gradients or high-frequency details) rather than spatially coherent patterns. This phenomenon can be observed in Fig. \ref{error}(b), where only portions of features along the channels are erased.  The observation motivates our design to sacrifice a portion of image pixels while keeping others unaffected. In addition, the distorted pixels can be generated with semantic accuracy and consistency. 

\subsubsection{Spatial-Error-Concentration Packetization} 
 Based on the above observations, we partition the feature tensor $\bm{y}_{Q}$ into discrete parts by leveraging a patch window of dimensions $W_z \times H_z$. Specifically, for each spatial position $(i,j)$ in $\bm{y}_{Q}$, we extract a local patch denoted by $\bm{y}_Q({i:i+W_z, j:j+H_z, :})$ spanning all $C_y$ channels, resulting in a sub-tensor of size $W_z \times H_z \times C_y$. These patches are then serialized into packets by grouping features along the channel axis ($C_y$), such that each packet $\bm{z}_{ij}$ corresponds to a spatially localized but channel-complete block, i.e., 
\begin{align}
	\bm{z}_{ij}
	&= \Large[ \bm{y}_Q({i:i+W_z, j:j+H_z, 1}), \dots, \nonumber \\
	& \quad \quad  \bm{y}_Q({i:i+W_z, j:j+H_z, C_y}) \Large] \\ & \quad \in \mathcal{U}^{W_z \times H_z \times C_y}.\nonumber
\end{align}
Then, $\{\bm{z}_{ij}\}$ is sequentially encoded into bit streams $\{\bm{b}_{1},\cdots,\bm{b}_{N}\}$.  Each bit stream $\bm{b}_{i}, i=1,2,\cdots,N$ is subsequently channel-encoded using a $(K, D)$ block code, producing a codeword $\bm{c}_k$ for transmission.  The channel rate  is calculated as $
	R_c=\frac{K}{D}=\frac{W_zH_zC_yR}{D} $ and the number packets for each image is calculated as $N=\frac{W_yH_y}{W_zH_z}$. 

It is worth noting that \emph{the proposed packetization strategy enables spatially localized protection, ensuring that packet errors primarily affect specific feature channels within localized spatial regions, rather than corrupting global spatial structures. } While this design may not protect the  average image quality, it plays a crucial role in preserving semantically important content in SemCom, as further elaborated in Section IV.

\begin{figure*}[t]
	\normalsize
	\setlength{\abovecaptionskip}{+0.3cm}
	\setlength{\belowcaptionskip}{-0.1cm}
	\centering
	\includegraphics[width=0.9\linewidth]{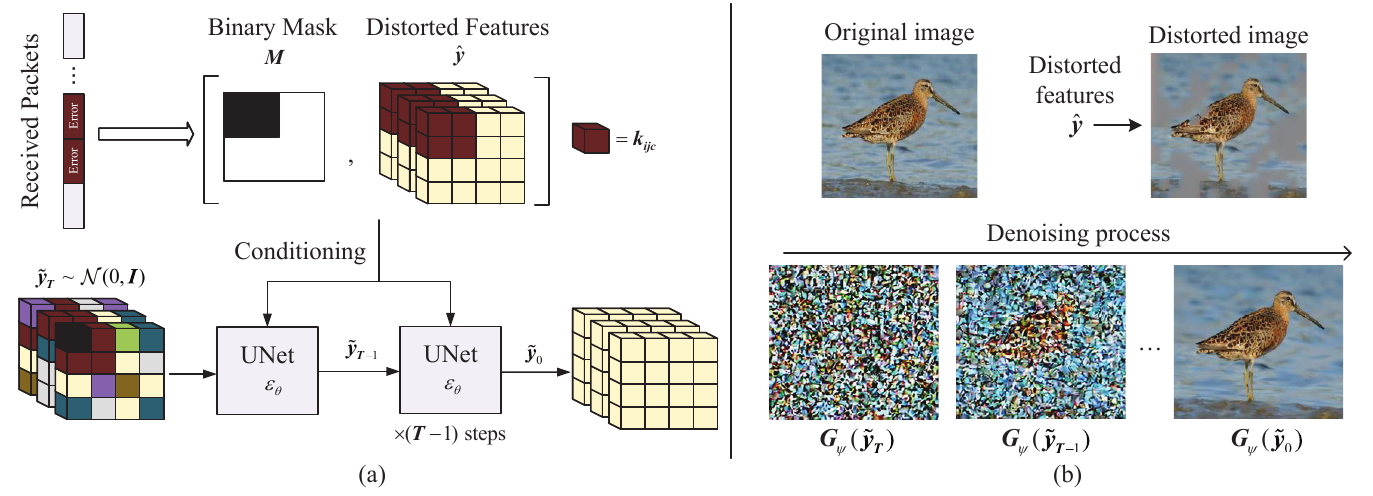}
	\captionsetup{justification=justified}
	\caption{Illustrations of the  diffusion-based feature imputing. Sub-figure (a) presents the framework of the generative process. Sub-figure (b) presents an image example to show the denoising process.  }
	\label{Neural_Arch2}
\end{figure*}

\subsection{Diffusion-based Feature Imputing}
In this subsection, we introduce the generative feature imputing method based on a diffusion model  to recover the erased features \cite{rombach2022high}.  As shown in Fig. \ref{Neural_Arch2}, the process details  are illustrated as follows.

\subsubsection{Construction of Attention Matrix $\texorpdfstring{\bm{M}}{M}$ and Feature $\hat{\texorpdfstring{\bm{y}}{y}}$}   According to the spatial-error-concentration packetization, the errors are concentrated on the spatial locations of features. After channel decoding,  the error position set $\mathbb{P}=\{(i,j,c)\}$ can be easily obtained. Based on the error set,  we construct a binary error matrix $\hat{\bm{M}} \in \{0,1\}^{W_y \times H_y}$  to indicate the spatial position of lost features. Its element follows: $\hat{M}_{ij} = 1$ if $ \forall (i,j,c) \in \mathbb{P}$ and $\hat{M}_{ij} = 0$ otherwise. To enhance spatial coherence, we propose to expand the attention region around each erased element by convolving $\hat{\bm{M}}$ with an all-ones kernel $\bm{\mathcal{K}}$ with the size of  $\kappa \times \kappa$, producing a perceptive mask $\bm{M}$:  
\begin{equation}
	\bm{M} = \mathbb{I}\left( \hat{\bm{M}} \ast \bm{\mathcal{K}} \geq 1 \right) \in \{0,1\}^{W_y\times H_y},
\end{equation}
where $\ast$ denotes convolution, and $\mathbb{I}[\cdot]$ denotes the element-wise indicator function, i.e., $M_{ij}=1$ if  $[\hat{\bm{M}} \ast \bm{\mathcal{K}}]_{ij}\geq1$  and $0$ otherwise. This dilation operation ensures that the diffusion model accounts for neighboring features when generating erased regions, mitigating abrupt transitions. The lost feature elements $\hat{y}_{ijc}, \forall (i,j,c) \in \mathbb{P}$ are initialized not with zeros but with  values $\{k_{ijc}\}$:  
\begin{equation}
\hat{y}_{ijc}=k_{ijc}=[F_{\phi}(\bm{O})]_{ijc}, \forall (i,j,c) \in \mathbb{P},
\end{equation}
where $\bm{O}$ is an all-zero matrix of dimension  $W\times H\times 3$. 

\subsubsection{Diffusion Process} Conditioned on the  constructed matrix $\bm{M}$ and feature $\hat{\bm{y}}$, the diffusion process recovers the lost features using the following two steps:
\begin{itemize}
	\item \textbf{Forward process}: Denote $\bm{y}_{0}$ as the initial feature of an image sample, i.e., $\bm{y}_0=F_{\psi}(\bm{x})$ (The quantization is not considered). At timestep $t$, Gaussian noise $\bm{\epsilon}_t$ is incrementally added to the  feature $\bm{y}_0$ according to a predefined schedule $\{\beta_t\}_{t=1}^T$ \cite{rombach2022high}:  
\begin{equation}
	\bm{y}_t | \bm{y}_{t-1} \sim \mathcal{N}\left(\sqrt{1 - \beta_t} \bm{y}_{t-1}, \beta_t \bm{I}\right).
\end{equation}
	In closed form, $\bm{y}_t$ can be directly expressed as:
	\begin{equation}
		\bm{y}_t = \sqrt{{\alpha}_t} \bm{y}_0 + \sqrt{1 - {\alpha}_t} \bm{\epsilon}_t, \quad \bm{\epsilon}_t \sim \mathcal{N}(0, \bm{I}),
	\end{equation}
	with ${\alpha}_t = \prod_{s=1}^t (1 - \beta_s)$.
	As time step goes, $\bm{y}_T$ will approximate isotropic Gaussian noise, i.e., $\bm{y}_{T} \sim \mathcal{N}(0,\bm{I})$.   
	\item \textbf{Reverse process}: In the reverse process, the features will be generated from the Gaussian noise $\mathcal{N}(0,\bm{I})$ step by step, as shown in Fig. \ref{Neural_Arch2} (a). Fig. \ref{Neural_Arch2} (b) visualizes the generation process given an image example. The UNet $\bm{\epsilon}_\theta(\cdot)$ with parameter $\theta$ is utilized to predict the noise $\bm{\epsilon}_t$ added to $\bm{y}_t$, conditioned on $\hat{\bm{y}}$, $\bm{M}$, and timestep $t$:  
	\begin{equation}
		\hat{\bm{\epsilon}}_{\theta,t}  = \bm{\epsilon}_\theta\left(\tilde{\bm{y}}_t, t,  \hat{\bm{y}},\bm{M}\right).
	\end{equation}
	The reverse process adopts the \emph{denoising diffusion implicit model} to progressively denoises $\tilde{\bm{y}}_t$ \cite{song2020denoising}:  
    \begin{align}
	\tilde{\bm{y}}_{t-1} &= \sqrt{\frac{\alpha_{t-1}}{\alpha_t}} \left( \tilde{\bm{y}}_t-\sqrt{(1-\alpha_t)}\hat{\bm{\epsilon}}_{\theta,t} \right) \nonumber \\
	& \quad + 
	\sqrt{1 - \alpha_{t-1} - \sigma_t^2} \hat{\bm{\epsilon}}_{\theta,t}  +
	\sigma_t  \bm{\iota},
\end{align}
	where $\sigma_t>0$ is the hyper-parameter, whose value decreases as the step goes,  and $\bm{\iota} \sim \mathcal{N}(\bm{0}, \bm{I})$.  
	After $T$ steps,  the generated feature $\tilde{\bm{y}}$ is output as $\tilde{\bm{y}}=\tilde{\bm{y}}_0$.
\end{itemize}

The UNet model is trained by minimizing the masked noise prediction error:  
\begin{equation}
	\mathcal{L}_{\text{diff}} = \mathbb{E}_{t, \bm{y}, \bm{\epsilon}_t} \left\| (1 - \bm{M}) \odot \left( \bm{\epsilon}_t - \bm{\epsilon}_\theta(\tilde{\bm{y}}_t, t,  \hat{\bm{y}},\bm{M}) \right) \right\|_2^2,
\end{equation}
where $\odot$ denotes the element-wise production.

\section{Advanced Technique -- Semantic-aware Power Allocation}
In this section, we propose a semantic-aware PA method  to protect the semantically important features against  channel fading.   First, we model the importance of features relevant to a specific task, such as classification, which also reflects the importance of the transmitted packets. Based on this importance modeling, we introduce the problem formulation and corresponding algorithms for PA.

\subsection{Semantically Importance Modeling}

We aim to quantify the importance of each spatial position in the feature map $\bm{y}$ related to the image label. To achieve this, we adopt the well-known class activation map (CAM) method \cite{zhou2016learning}  to generate the importance matrix $\bm{I} \in[0,1]^{W_y\times H_y}$.  The CAM method utilizes the inherent localization capabilities of CNNs \cite{zhou2016learning}: when a CNN-based architecture (e.g., ResNet-50) is trained for image classification, its backbone network (layers preceding the final fully-connected layer) implicitly learns to identify spatially discriminative features critical for class prediction. By leveraging this spatial property of features, we can directly utilize the pretrained backbone model to obtain the importance matrix without joint training. 

Specifically, denote the backbone model (e.g., ResNet-50) as $P_{\zeta}: \mathbb{R}^{W\times H \times 3} \rightarrow \mathbb{R}^{W_f \times H_f \times C_f}$ with $\zeta$ being the NN parameters. By inputting the image $\bm{x}$ into the backbone model, the latent feature can be calculated as
\begin{align}
	\bm{f}=P_{\zeta}(\bm{x}).
\end{align}
 Let $\bm{f}_c \in\mathbb{R}^{W_f\times H_f}, c=1,2,\cdots,C_f,$ denote the $c$-th feature map of $\bm{f}$.  When performing a  classification task, the extracted  feature $\bm{f}$ will be fed into a fully-connected neural network with a softmax function to output the probability of each label.  Specifically, the probability of label $k$, denoted by $P_{k}$, is calculated as 
\begin{align}
	P_k=\frac{e^{S_k}}{\sum_{k}e^{S_k}},  \text{with}\ S_k= \sum_{c}u_{c}^{k}\sum_{i,j} f_{ijc},
\end{align}
where $f_{ijc}$ denotes the $(i,j,c)$-th element of feature $\bm{f}$ and $u_{c}^k$ represents the parameter of the fully-connected neural network for label $k$. Essentially, $u_{c}^k$ indicates the importance of $c$-th feature map  for class $k$.  Hence, the importance matrix of class $k$ is calculated as a weighted  summation of the feature maps, i.e.,
\begin{align}
	\bm{I}^k = \sum_{c} u_{c}^k \bm{f}_c.
\end{align}
Then, each  element of $ \bm{I}^k$ is normalized into the interval $[0,1]$. 
Finally,  the normalized matrix  $\bm{I}^{k}$ is up-sampled to the spatial size of the feature $\bm{y}$, i.e. $(W_y,H_y)$.

The CAM process outputs the  importance matrices $\{\bm{I}^k\}$ of all the possible object labels. 
In this paper, we take one-label image as an example. Hence, we choose the importance matrix with the largest detection probability\footnote{The importance modeling process ensures that the transmitter has a coarse classification of the image. Even though the accuracy might not be high due to the simple architecture of CNNs, the importance matrix still captures the most discriminative object in the image, which enables the receiver for fine-grained vision tasks.}, i.e.,
\begin{align}
	\bm{I}=\bm{I}^k, \text{~with~} k=\arg \max ~P_k.
\end{align}
According to the packetization strategy proposed in Section III, the importance index, $w_{ij}$, for feature patch $\bm{z}_{ij}$ is averaged over the $W_z\times H_z$ grids, i.e.,
\begin{align} \label{importance_value}
	w_{ij}=&\frac{1}{W_zH_z} \sum_{k_1=i}^{i+W_z} \sum_{k_2=j}^{j+H_z} I_{k_1k_2},\forall  i=1,2,\cdots \frac{W_y}{W_z}, \nonumber \\
 &\forall j=1,2,\cdots, \frac{H_y}{H_z}. 
\end{align} 

Finally,  the importance indexes, $\{\omega_{k}, k=1,2,\cdots, N\}$, for transmission packets can be obtained by flattening the two-dimensional set $\{w_{ij}\}$ into a one-dimensional sequence according to the transmission order. 

\subsection{Problem Formulation}

For each packet duration time, we have $B$ packets allocated to $B$ frequency sub-carriers for transmissions. 
 Based on the importance modeling of these packets, our purpose is to optimize the PA to minimize the weighted sum of packet errors during each packet time for UEP.  For simplicity, denote $w_i, i=1,2,\cdots, B,$ derived in \eqref{importance_value}, as the importance index of packet $i$. Then, the optimization problem can be formulated as:
\begin{align} \label{pro:1}
	&\ \min_{\{P_i\}} \quad \sum_{i=1}^{B} \omega_i \rho_{i}(P_i)\\
	&\ \ \ \ \text{s.t.} \quad \frac{1}{B}\sum_{i=1}^{B}P_{i} \leq P_{ave},\nonumber \\
	&\ \ \ \  \quad \quad  P_{i}\geq 0, i=1,2,\cdots, B. \nonumber
\end{align} 
Problem \eqref{pro:1} is non-convex and hard to solve due to the channel dispersion term in \eqref{disper}.  To simplify the expression, the channel dispersion $V(\gamma_i)$  is   bounded  as a constant, i.e., $V (\gamma_i)\leq  1$, which also results in an upper bound of  $\rho_i(P_i)$. 
Then, we have 
\begin{align} \label{equ:bb}
	\rho_{i}(P_i) \leq  Q\left(\sqrt{D}(C(\gamma_i)-R_c) \log 2 \right) \triangleq \hat{\rho}_i(P_i).
\end{align}
Note that \eqref{equ:bb} holds if $R_c\leq C(\gamma_i)$. When $C(\gamma_i)>R_c$, $V=1$ is a good approximation as the error probability is too high to be of interest for reliable transmissions. 
Accordingly, the suboptimal solutions to Problem \eqref{pro:1}  can be  obtained by optimizing the following problem: 
\begin{align} \label{pro:2}
	&\ \min_{\{P_i\}} \quad \sum_{i=1}^{B} \omega_i \hat{\rho}_i(P_i)\\
	&\ \ \ \ \text{s.t.} \quad \frac{1}{B}\sum_{i=1}^{B}P_{i} \leq P_{ave},\nonumber \\
	&\ \ \ \  \quad \quad  P_{i}\geq 0, i=1,2,\cdots, B. \nonumber
\end{align} 

It is easy to check that the optimal PA must satisfy the maximum power budget, i.e., $\frac{1}{B}\sum_{i=1}^{B}P_{i} = P_{ave}$. Then, Problem \eqref{pro:2} equals to 
\begin{align} \label{pro:3}
	&\ \min_{\{P_i\}} \quad  \ \sum_{i=1}^{B} \omega_i \hat{\rho}_i(P_i) \\
	&\ \ \ \ \text{s.t.} \quad \frac{1}{B}\sum_{i=1}^{B}P_{i} = P_{ave},\nonumber \\
	&\ \ \ \  \quad \quad  P_{i}\geq 0, i=1,2,\cdots, B. \nonumber
\end{align} 

By calculating the second-order gradients of the objective function, it is easy to check that Problem \eqref{pro:3} is a convex problem for the region of $R_c\leq C(\gamma_i)$.  But it still remains non-convex for the region of $R_c> C(\gamma_i)$, making the whole problem difficult to solve.

\subsection{Power Allocation Algorithm}
This subsection presents an efficient algorithm to solve Problem \eqref{pro:3}.  To facilitate analysis, we obtain the following result.
\begin{proposition}\label{Lemma:1}
The sub-function $\hat{\rho}_i(P_i)$ in Problem \eqref{pro:3} can be upper bounded by
	\begin{align}\label{assum:3}
	\hat{\rho}_i(P_i) \leq \mathcal{G}(P_i)=		\left\{\begin{array}{ll}
		 	\hat{\rho}_i(P_i) & \text{\emph{if}}\ \ R_c \leq C(\gamma_i),\\
		\mu P_i+\tau, & \text{\emph{Otherwise}}.		
			\end{array}\right.
	\end{align}
where $\mu=-\sqrt{\frac{D}{{2\pi}}}\frac{|h_i|^2}{2^{R_c} \sigma^2}$ and $\tau=\frac{1}{2}+\sqrt{\frac{D}{{2\pi}}}\frac{2^{R_c}-1}{2^{R_c}}$. 
\end{proposition}
	\begin{IEEEproof}
	See Appendix A. 	
\end{IEEEproof}

By applying Proposition \ref{Lemma:1}, Problem \eqref{pro:3}  becomes
\begin{align} \label{pro:4}
		&\ \min_{\{P_i\}} \quad  \ \sum_{i=1}^{B} \omega_i \mathcal{G}(P_i) \\
	&\ \ \ \ \text{s.t.} \quad \frac{1}{B}\sum_{i=1}^{B}P_{i} = P_{ave},\nonumber \\
	&\ \ \ \  \quad \quad  P_{i}\geq 0, i=1,2,\cdots, B. \nonumber.
\end{align}
Problem \eqref{pro:4} is a convex problem and can be efficiently solved by using the Alternating Direction Method of Multipliers (ADMM) algorithm  \cite{boyd2004}. Note that theoretically, solving Problem \eqref{pro:4} leads to a suboptimal solution for the original problem in \eqref{pro:3}. However, the 
the upper bound in \eqref{assum:3} only affects the set $R_c>C(\gamma_i)$. When $R_c > C(\gamma_i)$, the transmission over this subcarrier becomes unreliable, leading to few power  in this region for both Problem \eqref{pro:3} and Problem \eqref{pro:4}. The other solutions are  obtained in the regime of $R_c \leq C(\gamma_i)$, which are not affected by the upper bound. Hence, the solution to Problem \eqref{pro:4} is very close to the one of Problem \eqref{pro:3}.


 \begin{figure*}[t]
	\normalsize
	\setlength{\abovecaptionskip}{+0.3cm}
	\setlength{\belowcaptionskip}{-0.1cm}
	\centering
	\includegraphics[width=0.7\linewidth]{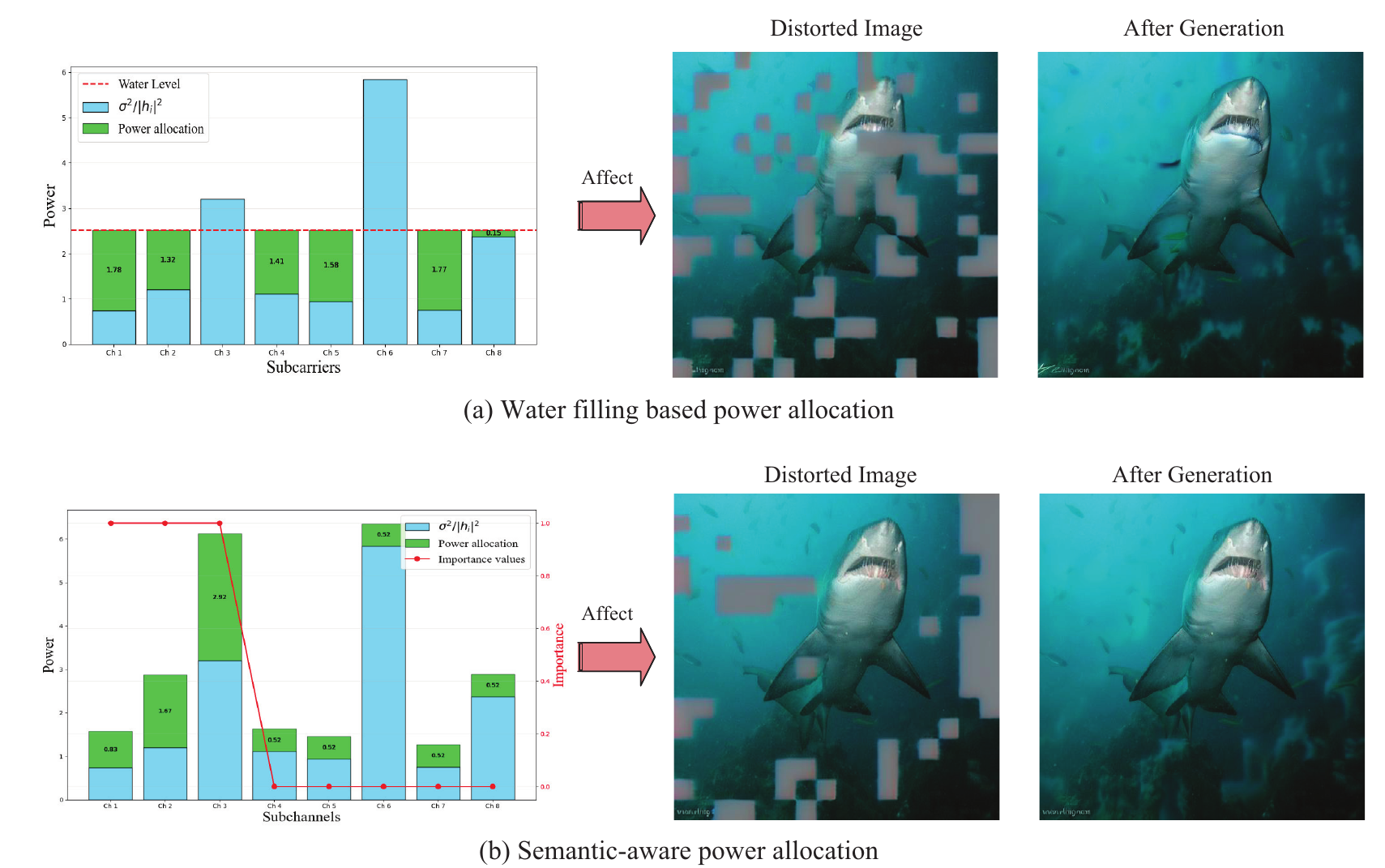}
	\captionsetup{justification=justified}
	\caption{Performance illustrations of the error-resilient SemCom with different PA schemes. The average packet error probability is around $0.2$. }
	\label{power}
\end{figure*}

%

\subsection{Discussions}
In this subsection, we discuss the effect of PA on the error-resilient SemCom system. For better illustrations, we analyze the transmission of a single image  at a SNR of $0$ dB.  As shown in Fig. \ref{power},  the reconstructed outputs of the distorted image  $G_{\psi}(\hat{\bm{y}})$ and the generated image $G_{\psi}(\tilde{\bm{y}})$  are compared under two PA strategies:
\begin{enumerate}
	\item \emph{Water-filling PA}: The traditional  water filling algorithm focuses on concentrating transmit power on subchannels with favorable channel conditions while neglecting weaker ones.  While considering limit transmission time \footnote{For fair comparisons, we consider that the transmission time for both the PAs are same. Since the water-filling PA disregards some poor subcarriers, the untransmitted features over the transmission time are regarded as lost. }, it risks packet losses in channels with poor conditions, regardless of the semantic importance of data. As shown in Fig.~\ref{power}(a), this results in \emph{random spatial errors} across the image. Critical semantic regions (e.g., the ``Shark'' object) suffer severe distortions, degrading the interpretability of the recovered image.
	\item \emph{Semantic-aware PA}: Our proposed strategy dynamically allocates power to the packets with high semantic importance, even in moderately poor channels. This ensures robust transmission of features critical to semantic accuracy, while allowing controlled losses in less important regions (e.g., image edges or backgrounds). As illustrated in Fig.~\ref{power}(b), errors are concentrated in non-critical areas, enabling the diffusion model to plausibly inpaint missing regions. The final output $G_\psi(\tilde{\bm{y}})$ retains key semantic details (e.g., the shark’s shape and texture), demonstrating superior alignment with the original semantic object.  However, it is worth noticing that the semantic-aware PA will sacrifice   subcarriers with good channel conditions (e.g. subcarrier 4,5 in Fig. \ref{power}) when the importance weights are low. As a result, even though this approach will maintain the semantic meaning of original images, more noncritical pixels might be distorted, leading to a low pixel-level performance.
\end{enumerate}

\section{Experimental Results}

\subsection{Experimental Settings}

\begin{enumerate}
	\item \textbf{Model Architecture}: In the experiments, we apply  the pretrained variational autoencoder (VAE)  from the latent diffusion framework \cite{rombach2022high} as the architectures of the semantic encoder and decoder. The uniform quantization is directly applied to the extracted latent features without retraining the encoder and decoder. For feature generation at the receiver, we employ the stable latent diffusion model \cite{rombach2022high}, leveraging its iterative denoising process to reconstruct lost features. To model semantic importance, we utilize a pretrained ResNet-50 backbone model, truncated before the final fully-connected layer, to generate CAMs. This ResNet-50 model is trained on the ImageNet dataset for the image classification task. 
	\item \textbf{Dataset}:  We evaluate our method on the widely adopted ImageNet-1k validation dataset \cite{image}, which contains high-resolution images with $1000$ object classes. All the images are resized as $512\times512$ pixels. 
	
	\item \textbf{Parameter Settings}:  For the finite blocklength transmissions, we set the block length $D=1024$, the channel rate $R_c=0.4375$, and the subcarrier number as $B=16$.  For the feature packetization, we set the size of patch windows as $(W_z,H_z)=(4,4)$. The kernel size for constructing binary matrix $\bm{M}$ is set as $\kappa=2$. The steps for diffusion-based generation is set as $T=10$.  The channel coefficient $h_i$ for each packet is determined based on the experimental settings: it is either set to $1$ for an AWGN channel, or sampled from a Rayleigh distribution, i.e., $h_{i} \sim \mathcal{CN}(0,1)$.  The quantization bits for each feature is set as $R=7$. The bits per pixel (Bpp) is calculated as $0.4375$. The channel uses per image element is calculated as $\frac{\text{Bpp}}{3R_c}=1/3$.  
	
	\item \textbf{Benchmarking Schemes}: We compare the proposed error-resilient SemCom with the  classic separate source-channel coding (SSCC) and DJSCC schemes:
	\begin{itemize}
		\item SSCC: For the SSCC scheme, we consider the traditional JPEG2000 and the VAE-based source coding methods \cite{Balle2018}, followed by the finite blocklength transmission. The former compresses images using  wavelet-based codec  and the later utilizes NN-based VAE to achieve higher rate-distortion efficiency. 
		\item DJSCC: The DJSCC baseline employs an end-to-end NN encoder-decoder architecture for robust image transmission \cite{kurka2020deepjscc}. Prior to decoding, channel equalization is applied to mitigate fading effects, and equal PA is adopted to ensure unit transmit power per symbol. 
	\end{itemize}
\end{enumerate}

\subsection{Evaluation of System Error Resilience}

 \begin{figure}[t]
	\centering
	\includegraphics[width=2.9in]{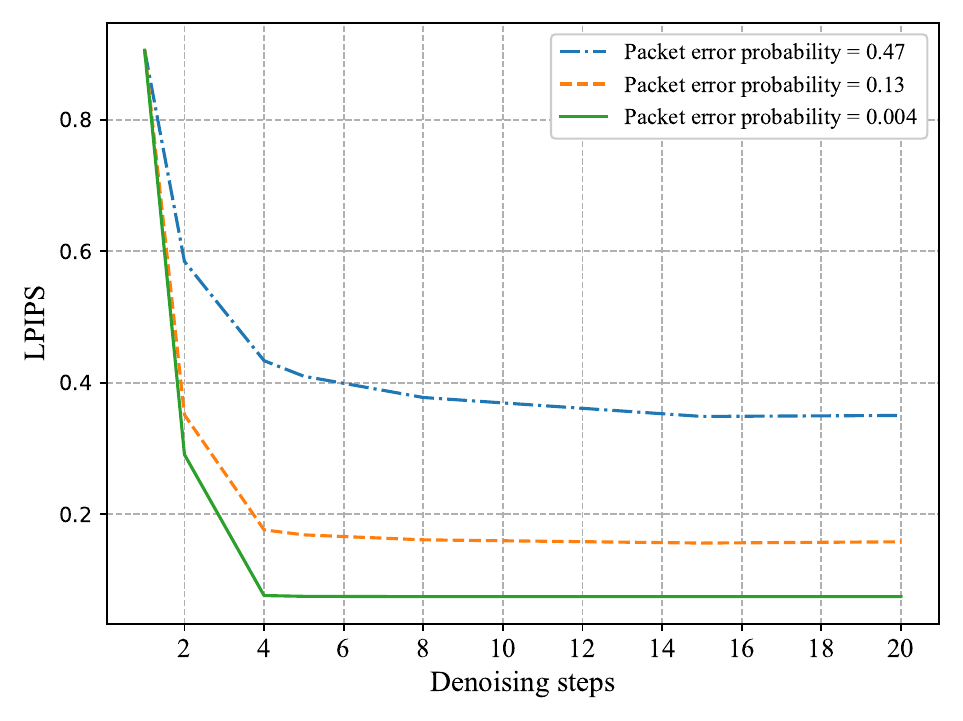}
	\captionsetup{justification=justified}
	\caption{Performance of the error-resilient SemCom versus  denoising steps.   }
	\label{conve}
\end{figure}

	\begin{figure*}[t]
	\normalsize
	\setlength{\abovecaptionskip}{+0.3cm}
	\setlength{\belowcaptionskip}{-0.1cm}
	\centering
	\subfigure[]{
		\begin{minipage}[t]{0.3\linewidth}
			\centering
			\includegraphics[width=2.2in]{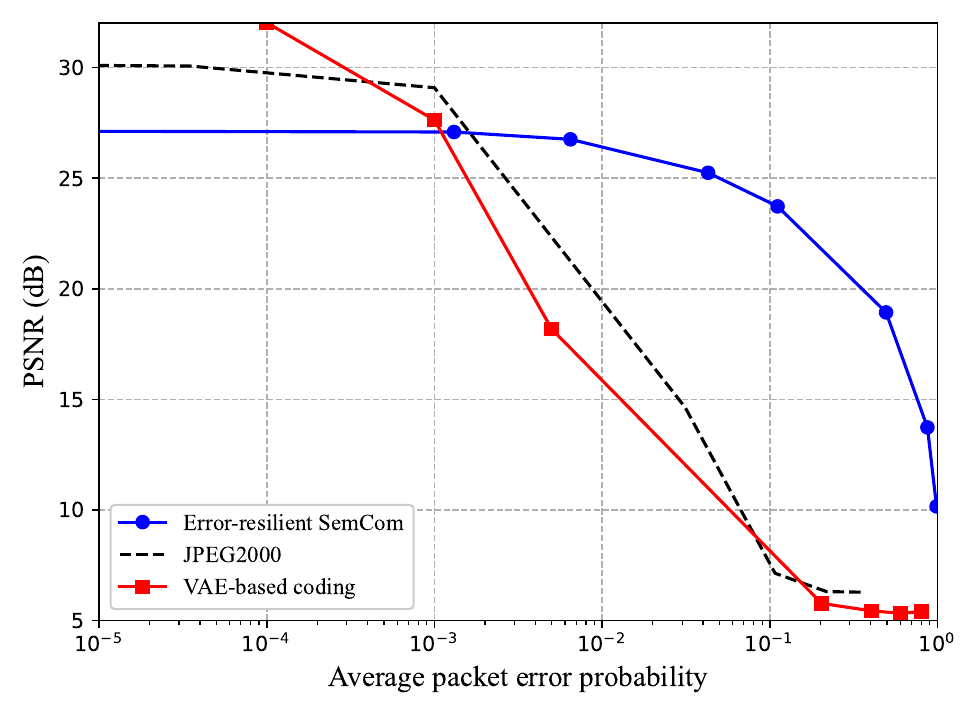}
			
		\end{minipage}%
	}%
	\subfigure[]{
		\begin{minipage}[t]{0.3\linewidth}
			\centering
			\includegraphics[width=2.2in]{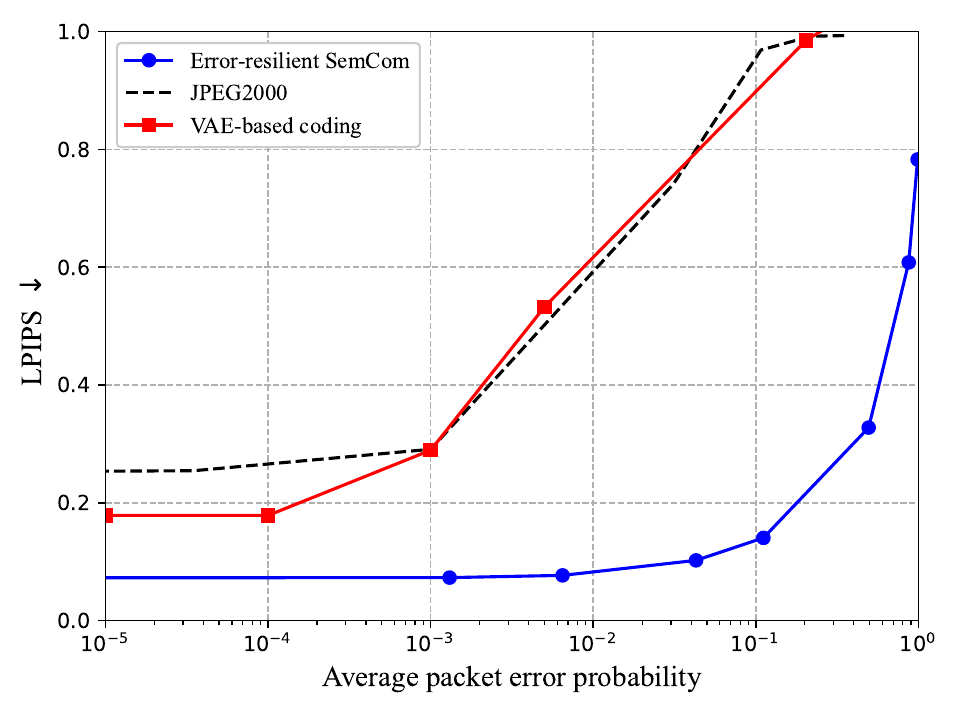}
			
		\end{minipage}%
	}%
	\subfigure[]{
		\begin{minipage}[t]{0.3\linewidth}
			\centering
			\includegraphics[width=2.2in]{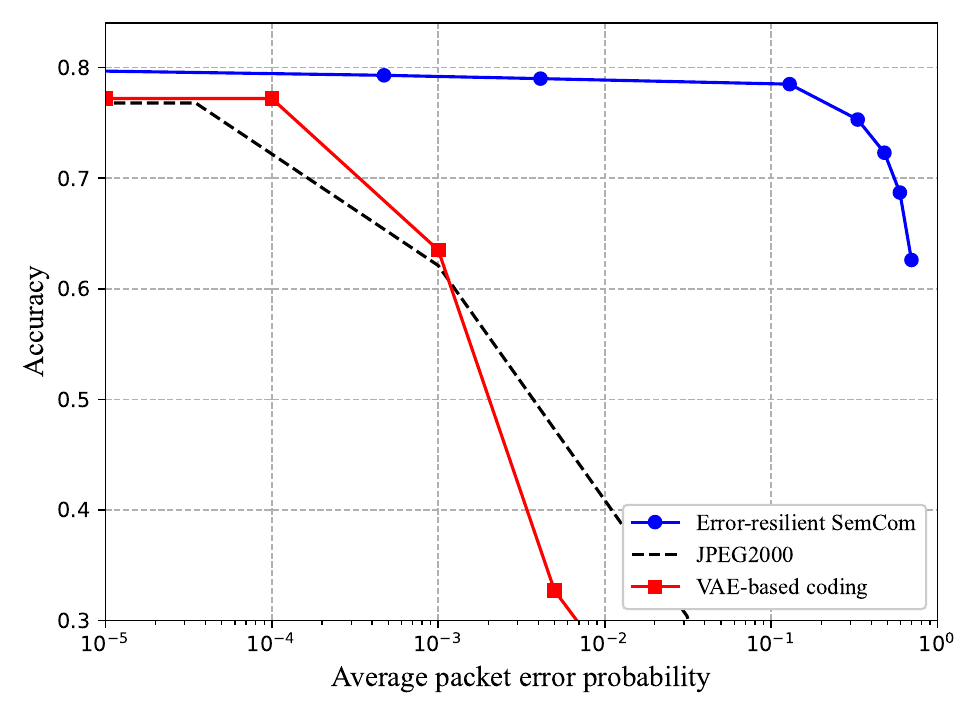}
		\end{minipage}%
	}%
	\captionsetup{justification=justified}
	\caption{Performance of the error-resilient SemCom under different packet error probabilities. }
	\label{fig:packeterror}
\end{figure*}

\subsubsection{Performance versus Denoising Steps} First, we investigate the performance of the error-resilient SemCom with different denoising steps. Fig. \ref{conve} plots the LPIPS metric as a function of the denoising steps. The results reveal that the proposed error-resilient SemCom only needs a few  denoising steps to achieve performance convergence, which is quite smaller than the traditional diffusion model needing hundreds steps. Notably, systems with higher packet error probabilities exhibit slower convergence rates. For example, a packet error probability of 
$0.47$ leads to convergence at step $16$, whereas a probability of 
$0.13$ achieves convergence by step $8$. This trend aligns with the expectation that greater channel distortion requires additional computational resource to mitigate its effects.

\subsubsection{Visual Performance Under Packet Errors}
Then, we  investigate the visual performance of the proposed error-resilient SemCom system under varying packet error probabilities. The tests are conducted over AWGN channel with equal PA. As depicted in Fig.~\ref{fig:packeterror} (a)-(b), the PSNR and LPIPS metrics are plotted as functions of the average packet error probability $\rho$. The results demonstrate that our system exhibits superior robustness to packet errors compared to both JPEG2000 and VAE-based separate coding schemes. For instance, at $\rho=0.1$, the proposed system achieves $\text{PSNR} = 23.7$~dB and $\text{LPIPS} = 0.14$, outperforming JPEG2000 ($7.1$~dB, $0.96$) and VAE-based coding ($7.2$~dB, $0.9$). Even at extreme high error probability, e.g., $\rho=0.5$, our scheme still can achieve high-fidelity image recovery with $\text{PSNR} = 19$~dB and $\text{LPIPS} = 0.32$. 
 This robustness stems from the proposed feature packetization and generation methods, which mitigate error propagation over image spatial domain and generate the lost features from knowledge. 

 However, at very low packet error probabilities ($\rho < 10^{-3}$), our scheme exhibits a  PSNR degradation  lower than JPEG2000 and VAE-based coding due to the absence of variable-length source coding. However,  maximizing the PSNR  only achieves the pixel-level accuracy, but it fails to account for the perception fidelity of images. In terms of LPIPS metric, which aligns better with human perception, our proposed scheme reveals consistent superiority compared with other schemes. At $\epsilon = 10^{-4}$, our system achieves around $0.12$ lower LPIPS compared with the VAE-based coding. 
 
 \subsubsection{Semantic Accuracy Under Packet Errors}
 
 Next,  we investigate the  accuracy  performance of the error-resilient SemCom under different packet error probabilities. Here, we utilize the semantic-aware PA to protect the semantic-critical features. The transformer-based classifier \cite{dosovitskiy2021image} is employed to detect the class label of the recovered image. As shown in Fig.~\ref{fig:packeterror} (c), it is observed that our proposed system reveals significant performance gain compared with other schemes. For example, our system maintains 79\% accuracy at a packet error probability of $\rho = 0.1$, outperforming conventional SSCC schemes by $50\%$ accuracy. At extreme high error probability $\rho=0.48$, the accuracy of the proposed scheme maintains around $72\%$. This robustness stems from the semantic relevance within a single image: only a small portion of an image’s pixels (e.g., foreground objects, edges) significantly contribute to its semantic meaning. By adaptively allocating power to the packets corresponding to these critical features, the proposed scheme can effectively preserve the image's semantic integrity.
 In contrast, JPEG2000 and VAE-based schemes exhibit steep accuracy declines (e.g., $<30\%$ at $\rho = 10^{-1}$) due to the error propagation phenomenon. For instance, losing a single packet in JPEG2000’s wavelet coefficients corrupts spatially contiguous regions, which severely degrades the semantic accuracy.

%


 \begin{figure}[t]
		\centering
	\subfigure[]{
	\begin{minipage}[t]{\linewidth}
		\centering
		\includegraphics[width=2.8in]{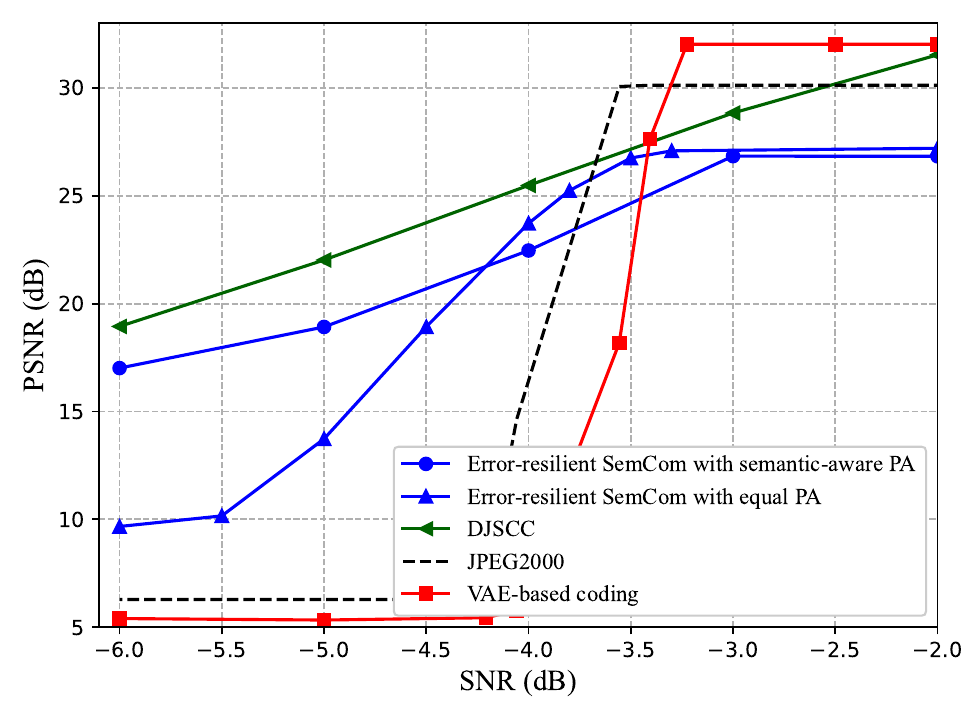}
		
	\end{minipage}%
}%

\subfigure[]{
	\begin{minipage}[t]{\linewidth}
		\centering
		\includegraphics[width=2.8in]{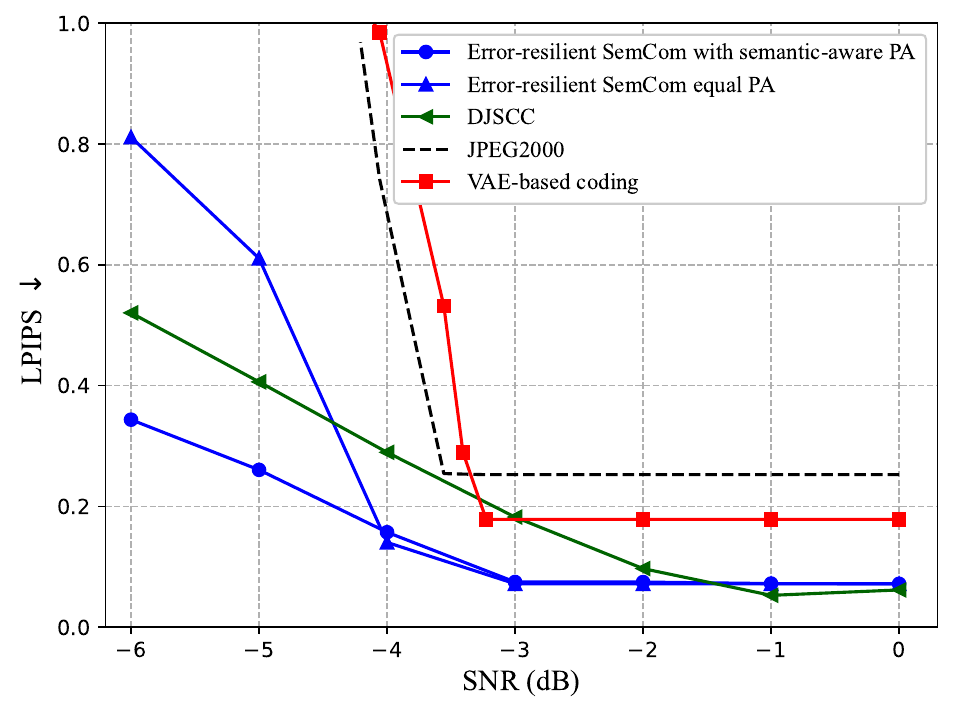}
	\end{minipage}%
}%
\captionsetup{justification=justified}
\caption{Visual performance of the error-resilient SemCom over AWGN channel. }
\label{AWGN}
\end{figure}

 \begin{figure}[t]
	\centering
	\includegraphics[width=2.8in]{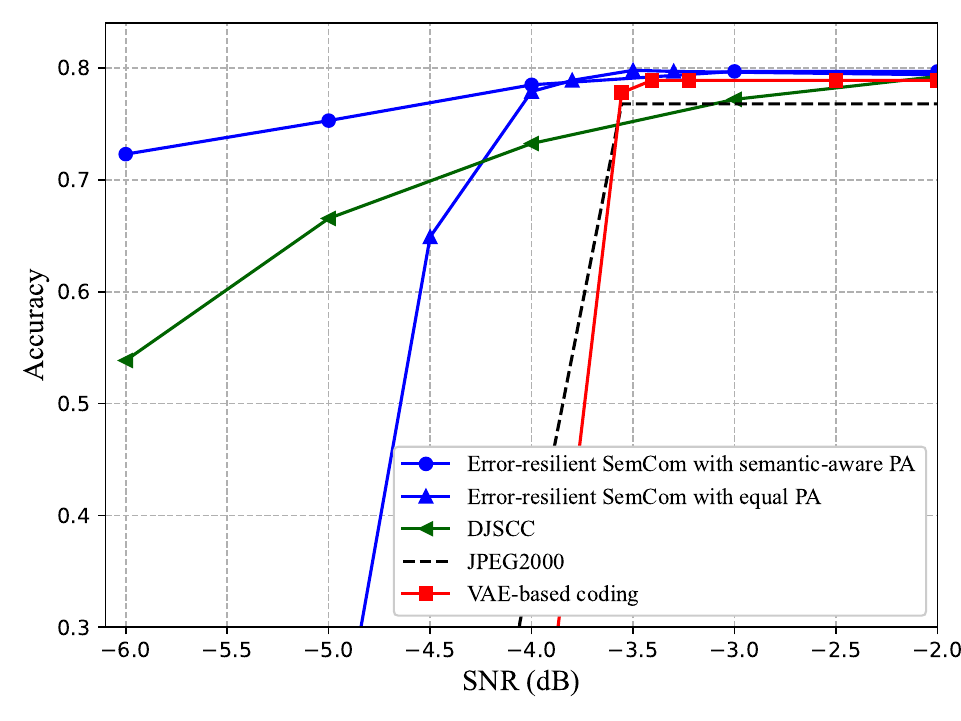}
	\captionsetup{justification=justified}
	\caption{Semantic accuracy of the error-resilient SemCom over AWGN channel.   }
	\label{accuracy_AWGN}
\end{figure}

\subsection{Performance Benchmarking  over AWGN Channel}

In this subsection, we compare  the error-resilient system with the other benchmark schemes over the AWGN channel. Since the channel coefficients are set as one for all the blocks, the water-filling PA equals to equal PA.  The DJSCC is trained at an SNR of $0$ dB and tested at other SNR regions. The channel uses per element are set as $1/3$ and are the same for all the schemes. 
Here, we consider the average SNR over blocks, which is defined as $\frac{P_{ave}}{\sigma^2}$. 

\subsubsection{Visual Performance} As shown in Fig.~\ref{AWGN}, we plot the PSNR and LPIPS metrics as functions of the channel SNR. It is observed that traditional SSCC schemes exhibit significant performance degradation as the channel SNR decreases. This phenomenon, termed the \emph{cliff effect}, arises primarily from error propagation in SSCC pipelines. For instance, JPEG2000 experiences a catastrophic $25$ dB PSNR drop between $\text{SNR} = -4.2$~dB and $\text{SNR} = -3.7$~dB, where channel noise disrupts its entropy-coded bitstream. In contrast, our error-resilient SemCom system demonstrates graceful degradation, similar to analog DJSCC. At $\text{SNR} = -4.5$~dB, our scheme achieves a $12$ dB PSNR improvement over JPEG2000 and VAE-based schemes. Furthermore, the proposed system achieves superior LPIPS performance compared to DJSCC: to attain an LPIPS of $0.2$, our semantic-aware PA scheme requires $1.5$ dB less SNR than DJSCC.  

An interesting  observation is the semantic-aware PA’s ability to preserve high visual quality at low SNRs compared to equal PA. For example, at $\text{SNR} = -6$~dB, semantic-aware PA achieves a $7$ dB PSNR gain over equal PA. This stems from its focus on protecting semantic-critical features, which dominate perceptual quality. However, at higher SNRs (e.g., $-4$~dB to $-3$~dB), equal PA slightly outperforms semantic-aware PA by $1 \sim 2$ dB PSNR. This occurs because both schemes preserve critical features in favorable channel conditions, but semantic-aware PA sacrifices power allocated to non-critical regions (e.g.,  backgrounds), slightly reducing pixel-level fidelity.

\subsubsection{Semantic Accuracy}Fig.~\ref{accuracy_AWGN} plots the classification accuracy as a function of the channel SNR. The proposed error-resilient SemCom system with semantic-aware PA achieves superior accuracy compared to both DJSCC and SSCC baselines. For instance, to attain an accuracy of $0.72$, the semantic-aware PA scheme requires $\sim$$1.5$~dB and $2.3$~dB less SNR than DJSCC and SSCC, respectively. This performance gain aligns with the empirical observation that only a small subset of features, such as object, spatial relationships, are critical to an image’s semantic meaning. By dynamically allocating resources to protect these features, our system avoids wasting resources on semantically redundant regions (e.g., uniform backgrounds), thereby maintaining robustness even at extremely poor channel conditions.

\subsection{Performance Benchmarking over Rayleigh Fading Channel}
In this subsection, we study the performance of the error-resilient SemCom system over a Rayleigh fading channel. The DJSCC is trained at an SNR of $10$ dB.

\begin{figure}[t]
	\centering
	\subfigure[]{
		\begin{minipage}[t]{\linewidth}
			\centering
			\includegraphics[width=2.8in]{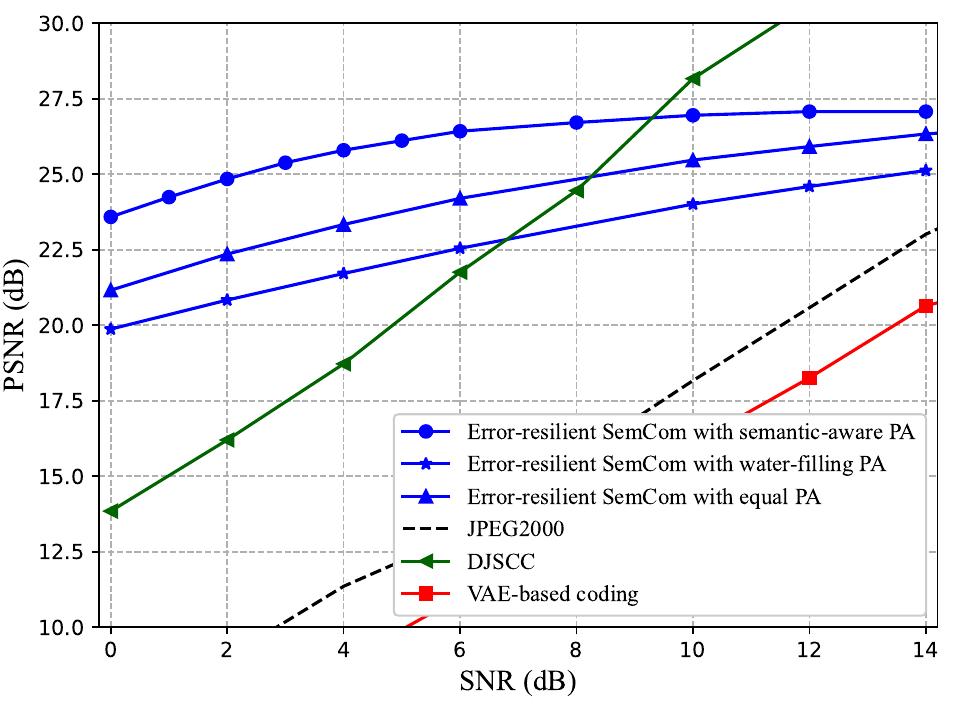}
			
		\end{minipage}%
	}%
	
	\subfigure[]{
		\begin{minipage}[t]{\linewidth}
			\centering
			\includegraphics[width=2.8in]{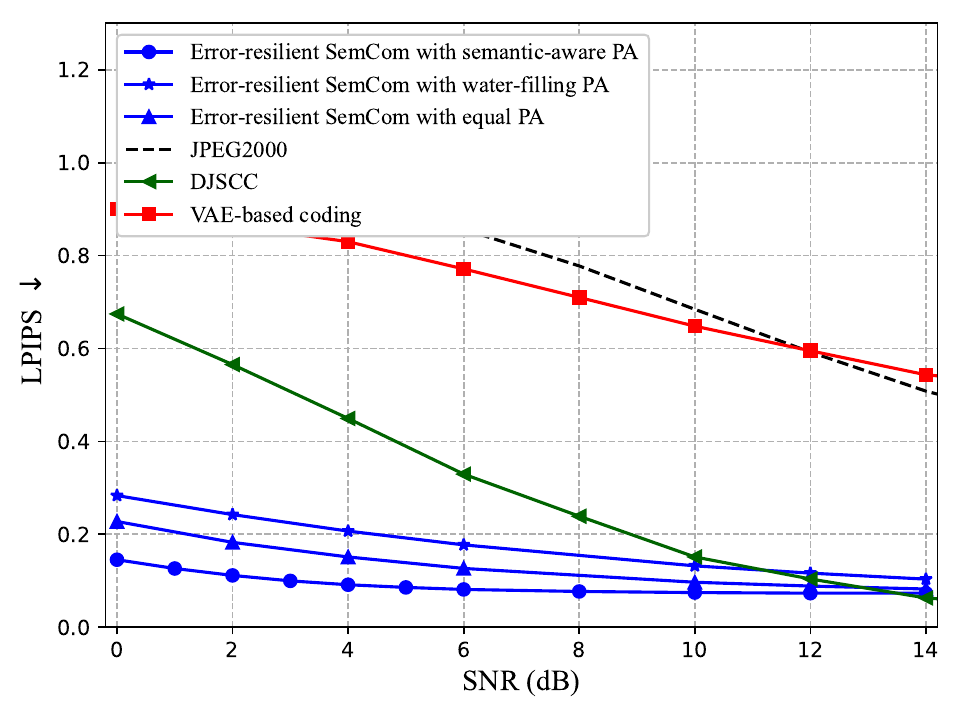}
			
		\end{minipage}%
	}%
	\captionsetup{justification=justified}
	\caption{Visual performance of the error-resilient SemCom over block Rayleigh fading channel. }
	\label{fading}
\end{figure}

\begin{figure}[t]
	\centering
	\includegraphics[width=2.8in]{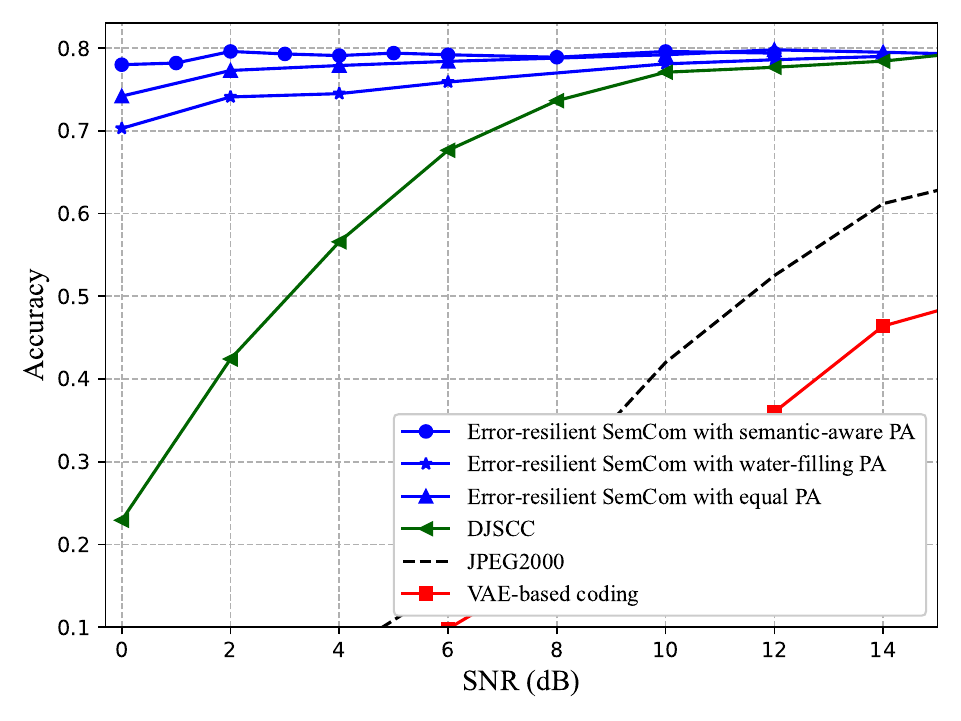}
	\captionsetup{justification=justified}
	\caption{Semantic accuracy of the error-resilient SemCom over block Rayleigh fading channel.   }
	\label{accruacy_fading}
\end{figure}

\subsubsection{Visual Performance}  
\begin{figure*}[t]
	\normalsize
	\setlength{\abovecaptionskip}{+0.3cm}
	\setlength{\belowcaptionskip}{-0.1cm}
	\centering
	\includegraphics[width=0.92\linewidth]{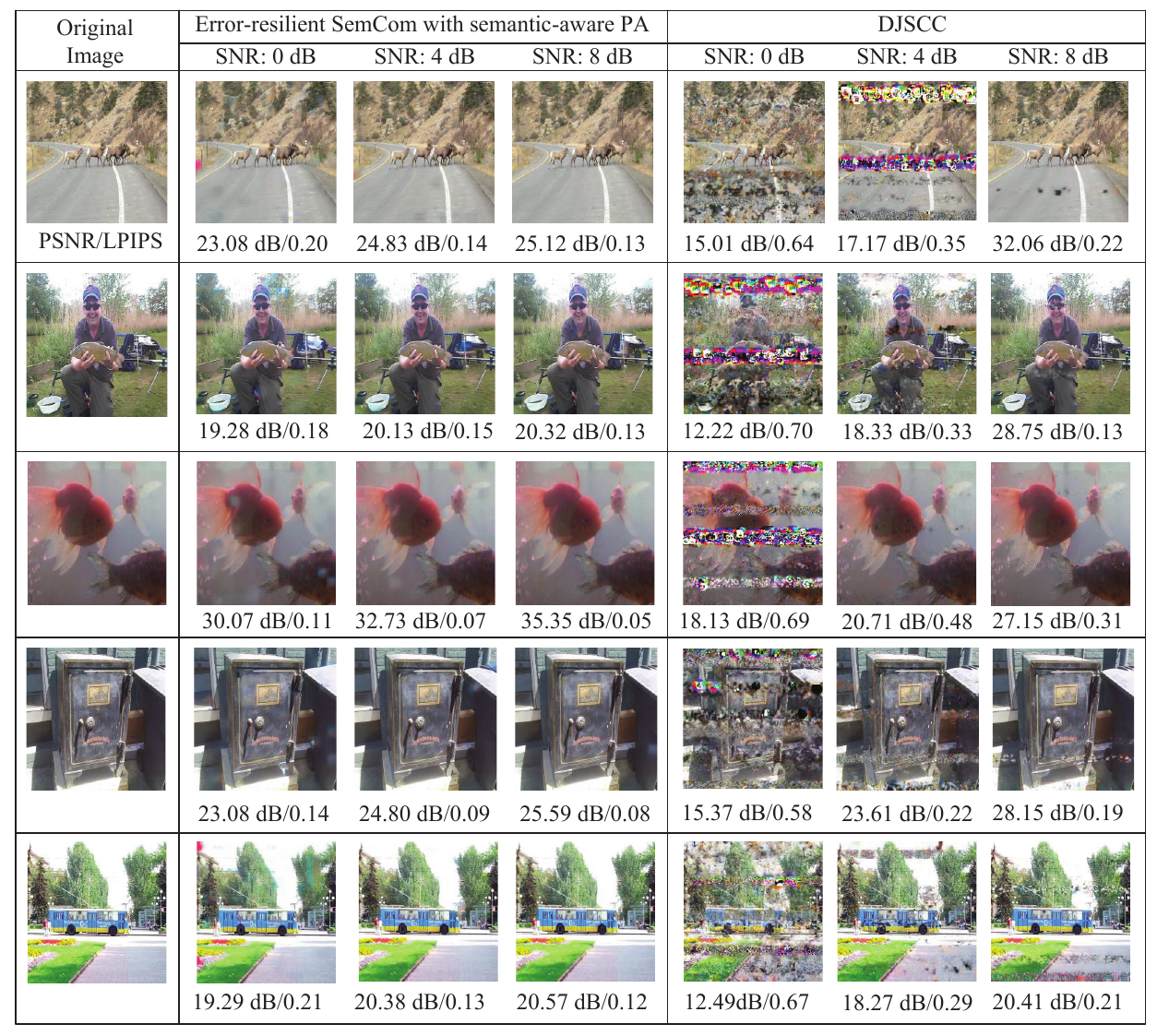}
	\captionsetup{justification=justified}
	\caption{Visualization examples of the error-resilient SemCom and DJSCC methods over block Rayleigh fading channel. The examples of VAE-based and JPEG2000 methods are ignored, since they cannot be successfully decoded under the presented SNR conditions.   }
	\label{example}
\end{figure*}

Fig.~\ref{fading} depicts the PSNR and LPIPS metrics as functions of average SNR over a block Rayleigh fading channel. Compared to the AWGN channel, deep fading introduces severe distortions in image recovery, necessitating higher SNR for comparable performance. While the DJSCC and SSCC schemes achieve  gains at high SNRs, the proposed error-resilient SemCom system exhibits superior robustness in low-SNR regimes. For instance, at $\text{SNR} = 0$~dB, our system with semantic-aware and equal PA achieves $11 $ dB and $8$ dB PSNR improvements over DJSCC, alongside a $0.4 $ LPIPS gain. The reason for this phenomenon  is that the DJSCC scheme without the packetization strategy is more sensitive to deep fading, as the distorted symbols may corrupt global image pixels. In contrast, our packetized SemCom localizes errors to non-critical features, preserving semantic fidelity even under deep fading.  
Visual examples in Fig.~\ref{example} substantiate this analysis. For the first image at $\text{SNR} = 0$~dB, our system exhibits different content in localized regions (e.g., road textures, tree outlines) but maintains high perceptual quality ($\text{LPIPS} = 0.2$). DJSCC, however, suffers from global distortions (e.g., blurred foregrounds, loss of object boundaries) due to the symbol corruption. Similar phenomena are observed across other test images.  

Another  observation in Fig.~\ref{fading} is that  the performance of the equal PA is always better than the water-filling PA. At $\text{SNR} = 0$~dB, equal PA achieves a $1 $~dB PSNR gain. This occurs because water-filling PA optimizes for maximizing Shannon capacity, which assumes asymptotically long blocklengths and is suboptimal for finite blocklength transmission. In finite blocklength transmissions,  packet errors are non-negligible, and water-filling’s aggressive PA to high-gain subcarriers starves critical semantic features in weaker subcarriers. Equal PA’s uniform power distribution mitigates this by ensuring baseline protection for all features, avoiding catastrophic semantic losses from disregarded packets.  

\subsubsection{Semantic Accuracy}   Fig.~\ref{accruacy_fading} evaluates the classification accuracy of the proposed system under block Rayleigh fading. It is observed that our system's robustness is particularly evident at low SNR regimes. The error-resilient SemCom with semantic-aware PA achieves above $79\%$ accuracy  at $\text{SNR} = 0$~dB, outperforming DJSCC ($25$\%) and SSCC ($<1\%$). To attain $70\%$ accuracy, our system requires $7$ dB less SNR than DJSCC.  In addition, it is observed that even though the PSNR metric of the DJSCC is high, the accuracy is lower than our proposed scheme. 

\section{Concluding Remarks}
In this paper, we have presented the framework of  generative feature imputing  for digital SemCom systems, aimed at improving robustness against packet errors and enhancing semantic fidelity. By introducing the spatial-error-concentration packetization strategy and the generative feature imputing method, the proposed approach spatially concentrates distortions and effectively mitigates the cliff effect inherent in digital transmissions. Furthermore, a semantic-aware PA  was developed to enable UEP based on semantic importance, further improving the fidelity of reconstructed images. Extensive experiments demonstrated that the proposed framework outperforms traditional methods such as DJSCC and VAE-based coding in terms of semantic accuracy and perceptual quality under low SNR conditions.

However, the current framework does not incorporate variable-length source coding mechanisms, which helps reduce coding latency but limits overall coding efficiency. Future work may extend the generative feature imputing framework into two directions: (1) developing more advanced semantic coding and GenAI-based imputing techniques to enhance efficiency and reconstruction quality; and (2) exploring its application to UEP-related tasks such as bandwidth allocation and channel rate control.

\appendices
\section{Proof of Proposition \ref{Lemma:1}}
Proposition \ref{Lemma:1} can be  obtained by proving $\hat{\rho}_i(P_i)\leq uP_i+\tau$, with $P_i$ satisfying $R_c> \log_2(1+\frac{|h|_i^2P_i}{\sigma^2})$. Let $g_i=\frac{|h_i|^2}{\sigma^2}$. The second-order gradient of sub-function $\hat{\rho}_i(P_i)$ in Problem \eqref{pro:3} is calculated as 
\begin{align}
	&\frac{\partial^2 \hat{\rho}_i(P_i)}{\partial P_i^2} \\ \nonumber 
	&= \frac{\sqrt{D}g_i^2}{\sqrt{2\pi}(1+g_iP_i)^2 } \exp\left(-\frac{D(\log(1+g_iP_i)- R_c \log2)^2}{2}\right) \\ \nonumber 
	&\quad \cdot \left[D(\log(1+g_iP_i)- R_c \log2)+1 \right]. 
\end{align}
It is easy to check that  $\frac{\partial^2 \hat{\rho}_i(P_i)}{\partial P_i^2} \geq 0$ if and only if $R_c \leq \log_2(1+g_iP_i)+\frac{1}{D}\log_2 e$. Since  the block length $D$ is sufficiently large (e.g., $256$, $512$, and  $1024$) in practical communication system, the term $\frac{1}{D}\log_2 e$ is small and can be omitted. Hence, we conclude that function $\hat{\rho}_i(P_i)$ is convex when $R_c \leq \log_2(1+g_iP_i)$ and is concave in the region of $R_c \geq \log_2(1+g_iP_i)$.  Let $\tilde{P}$ be the intersection of these two sets, i.e., $\tilde{P}=\frac{2^{R_c}-1}{g_i}$. 

According to the property of concave function \cite{boyd2004}, we have 
\begin{align} 
	\hat{\rho}_i(\hat{P})-\hat{\rho}_i(\tilde{P})\leq \frac{\partial \hat{\rho}_i(P_i)}{\partial P_i}|_{P_i=\tilde{P}} \cdot(\hat{P}-\tilde{P}), \forall 0 \leq \hat{P} \leq \tilde{P}.
\end{align}
which implies that 
\begin{align}
		\hat{\rho}_i(\hat{P}) \leq -\frac{\sqrt{D}g_i}{\sqrt{2\pi} 2^{R_c}}\hat{P}+\frac{\sqrt{D}(2^{R_c}-1)}{\sqrt{2\pi}2^{R_c}}+\frac{1}{2}. 
\end{align}
Hence, Proposition  \ref{Lemma:1} is proved. 

\bibliographystyle{IEEEtran}
\bibliography{semantic}

\end{document}